\documentclass{article} 
\usepackage{iclr2020_conference}

\usepackage{hyperref}
\usepackage{url}
\usepackage{times}
\usepackage{epsfig}
\usepackage{graphicx}
\usepackage{amsmath}
\usepackage{amssymb}
\usepackage{xcolor,colortbl}
\usepackage{times}
\usepackage[utf8]{inputenc}
\usepackage[small]{caption}
\usepackage{algorithm2e}
\usepackage{hhline}
\usepackage{multirow}
\usepackage{subfigure}

\newcommand{\specialcellm}{\multirow{2}{*}{\rotatebox[origin=c]{0}{\begin{tabular}[c]{@{}c@{}}Na\"ive\end{tabular}}}}
\newcommand{\specialcelln}{\multirow{2}{*}{\rotatebox[origin=c]{0}{\begin{tabular}[c]{@{}c@{}}Human\end{tabular}}}}

\usepackage[authormarkup=none]{changes}
\definechangesauthor[name={DanH},color=red]{Dan}

\title{Transfer Learning of Photometric Phenotypes in Agriculture Using Metadata}

\author{Dan Halbersberg$^1$ \& Aharon Bar Hillel$^1$ \& Shon Mendelson$^1$ \& Daniel Koster$^2$ \& Lena Karol$^2$\\
	\textbf{\& Boaz Lerner$^1$} \\
	$^1$ Department of Industrial Engineering and Management \\
	Ben-Gurion University of the Negev, Israel\\
	\texttt{\{halbersb,barhille,shonm,boaz\}@post.bgu.ac.il} \\
	$^2$ Hazera Ltd. Berurim M.P Shikmim, Israel \\
	\texttt{\{daniel.koster,Lena.Karol\}@hazera.com} \\
}

\iclrfinalcopy 

\begin{document}

\maketitle

\begin{abstract}
	Estimation of photometric plant phenotypes (e.g., hue, shine, chroma) in field conditions is important for decisions on the expected yield quality, fruit ripeness, and need for further breeding. Estimating these from images is difficult due to large variances in lighting conditions, shadows, and sensor properties. We combine the image and metadata regarding capturing conditions embedded into a network, enabling more accurate estimation and transfer between different conditions. Compared to a state-of-the-art deep CNN and a human expert, metadata embedding improves the estimation of the tomato’s hue and chroma.
\end{abstract}

\section{Introduction and materials}\label{Sec.Introduction}

Shine, color, and chroma are important photometric phenotypes to breeders during agricultural cultivation and to growers quantifying the quality of their yield~\citep{castellar2003color,motonaga2013fruit,udomkun2014laser} because these phenotypes stimulate buyers to consume and enjoy fruits~\citep{barrett2008color,yoshioka2010image}, help to determine harvest time during ripening~\citep{mejia2009effect,soltani2011changes}, and can be used to predict the color of the final product~\citep{barrett2008color}. Tomatoes, e.g., are known for their vibrant red color (due to the presence of carotenoid, lycopene, and beta-carotene), and those having a deeper red color are often more mature, sweeter, and contain more lycopenes~\citep{barrett2008color}. Here, we focus on hue and chroma, both of which are related to color perception. The former measures the combination of the fundamental (unique) colors as interpreted by our brain~\citep{naik2003hue}, and the latter measures how different from gray an object's color appears to be~\citep{fairchild2013color}.

\begin{figure}[hb!]
	\centering
	{
		\subfigure[]{\label{FigureCorr}
			\includegraphics[width=0.18\textwidth]{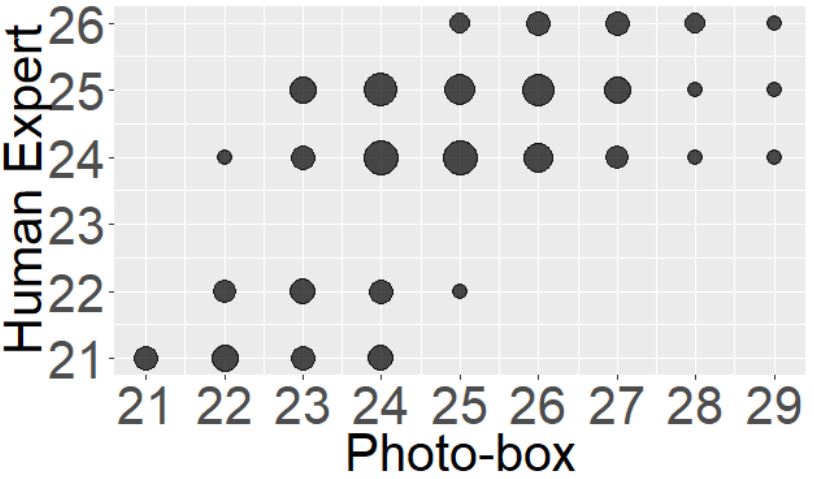}
		}
		\subfigure[]{\label{a}
			\includegraphics[width=0.15\textwidth]{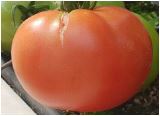}
		}
		\subfigure[]{\label{b}
			\includegraphics[width=0.15\textwidth]{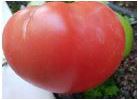}
		}
		\subfigure[]{\label{c}
			\includegraphics[width=0.14\textwidth]{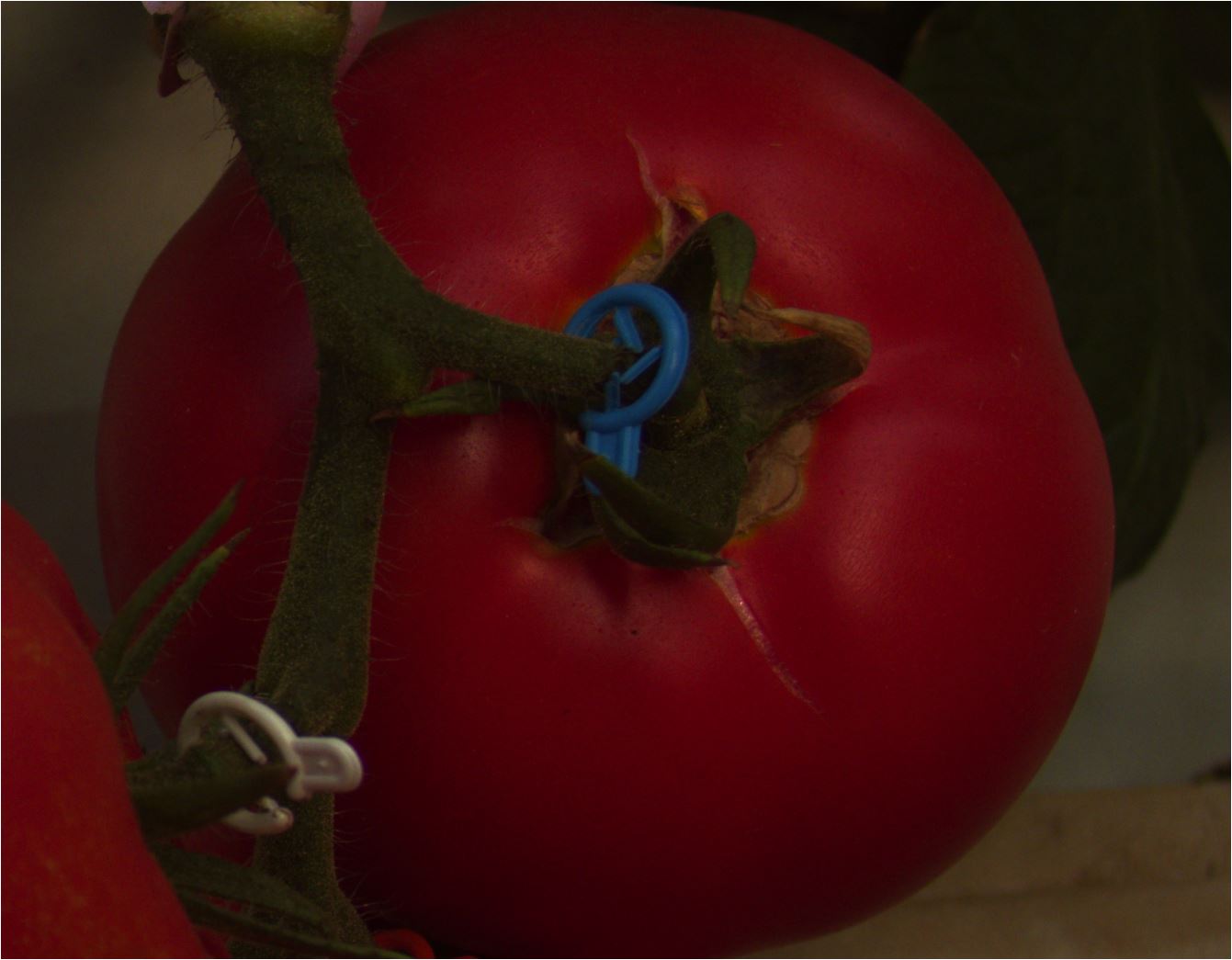}
		}
		\subfigure[]{\label{d}
			\includegraphics[width=0.12\textwidth]{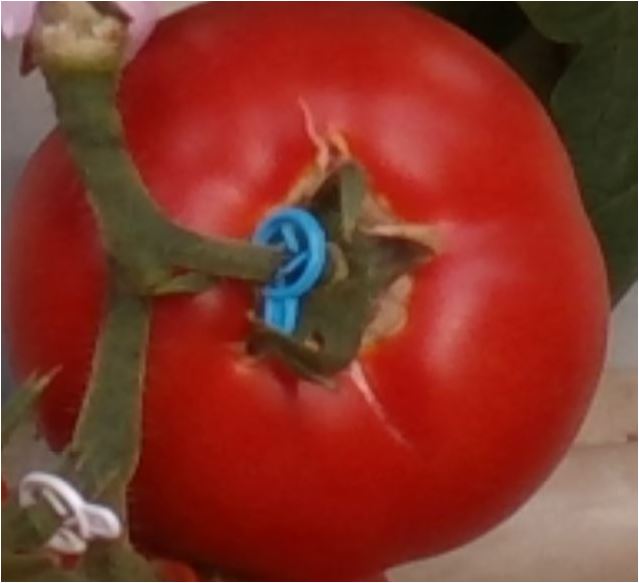}
		}
	}
	\caption{(a) Expert-grower estimation vs. photo-box hue values for $\approx200$ tomatoes reveal only 0.56 correlation. Images of the same tomato taken by the same camera on the same day at noon (b) and in the morning (c), and images of the same tomato taken on the same day and hour by two different cameras (d) and (e).}
	\label{FigureTomato}
\end{figure}

Measuring photometric phenotypes, e.g., hue and shine, in fruits manually by human experts is time-consuming, expensive, and subjective, raising doubts on its reliability~\citep{li2014review,yoshioka2010image}. In contrast, the use of instrumental measurements, e.g., Maxwell's spinning disk or a photo-box~\citep{barrett2008color} is fast, cheap, and objective (Fig.~\ref{FigureCorr}). But because phenotyping using a photo-box requires picking the fruit, it is limited to a small number of fruits.

A solution is to measure these phenotypes from images, and indeed, recently, several agricultural phenotyping tasks based on computer vision have been proposed, e.g., fruit/leaf counting, mutant or stress classification, and growth rate and age estimation~\citep{kamilaris2018deep,li2014review,rousseau2013high,song2014automatic,ubbens2017deep,vit2019length}. However, regarding photometric phenotyping, major confounding/intervening factors, e.g., environmental (and specifically lighting) conditions (e.g., season, part of the day, clouds) and the camera used, diversify such phenotyping in the image~\citep{nakazato2008environmental} (Fig.~\ref{a}--~\ref{d}).

We propose the \textit{metadata embedding for photometric phenotyping} (MEPP) method to cope with the influence of environmental conditions on photometric agricultural phenotypes by embedding metadata about the environment (e.g., lighting condition and camera characteristics) into a deep neural network (DNN). The MEPP enables (1) transfer learning over the exponentially increasing number of metadata variable (unseen/rare) combinations, and (2) standardization of fruit photometric phenotyping.~We demonstrate the MEPP in greenhouse tomato photometric phenotyping of hue and chroma.~Our contribution is threefold: (1) a metadata-image combined DNN architecture, (2) a method to measure photometric phenotypes from images which allows transfer learning between environmental conditions, and (3) demonstration of photometric phenotyping of tomato hue and chroma in agricultural images that is superior to a state-of-the-art DNN and the human expert.

We used a dataset collected in a tomato greenhouse of a global leader in the seed industry consisting of $\approx820$ images of $\approx200$ tomatoes. Each tomato was photographed on the same date in the morning, noon, and afternoon, using high- and low-resolution cameras (attached to each other to maintain the same acquisition conditions), and in direct sun and in shadow (most tomatoes could not be photographed in all conditions). Along the  experiment, we kept fixed the cameras' angles and exposure of the high resolution camera (that of the low resolution camera did not support that). The lighting condition was measured for each image using a light sensor (PixelSensor$^{TM}$) that was positioned on one of the cameras. After completing the experiment, and on the same day, the tomatoes were harvested and evaluated by a breeder expert and, independently, their hue and chroma were measured using a photo-box. Overall, each of the 820 single-object tomato images had photo-box-based labels for hue and chroma and was associated with the metadata variables: (1) \textit{Camera} (high/low resolution), (2) \textit{Part of Day} (morning/noon/afternoon), (3) \textit{Light sensor} (a continuous value), (4) \textit{Direct sun/shadow}, and (5) \textit{Plant type} (regular/cherry).

\section{The Proposed MEPP architecture}\label{SecArch}

Our proposed MEPP architecture to predict photometric phenotypes embeds metadata into a DNN, and thereby allows photometric phenotype prediction based on both image measurements and knowledge about the capture/environmental conditions. Embedding promotes transfer learning by which, although the number of metadata variable combinations increases exponentially with the number of variables, the MEPP can predict even for a combination unseen before.

\begin{figure*}[ht!]
	\scriptsize
	\centering
	\begin{minipage}[b!]{.55\textwidth}
		\centering	
		\includegraphics[width=0.95\textwidth]{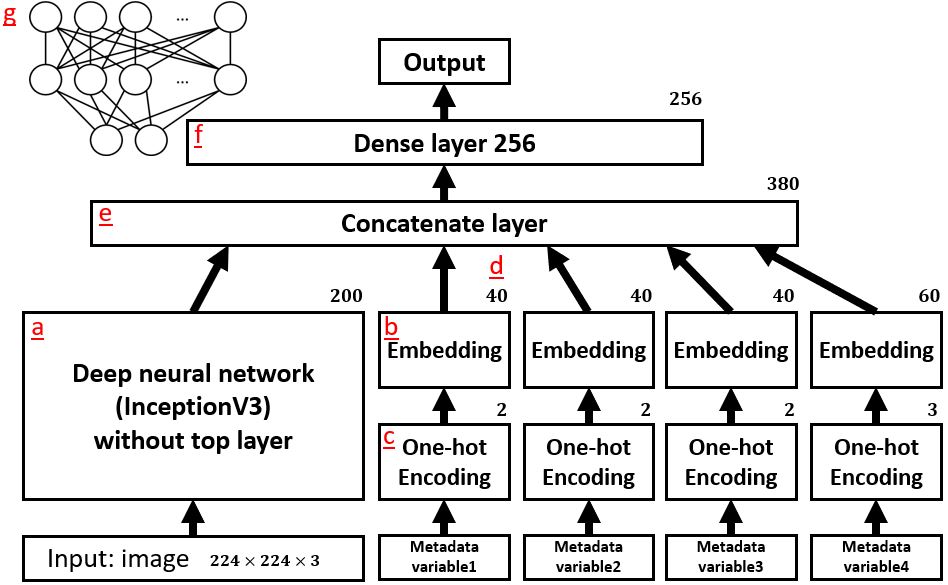}
		\captionof{subfigure}{Our proposed MEPP architecture.}
	\end{minipage}\hspace*{-0.75em}
	\begin{minipage}[t!]{.45\textwidth}
		\centering
		{
			\subfigure[Learning curves]{\label{Figurehue}
				\includegraphics[width=0.8\textwidth]{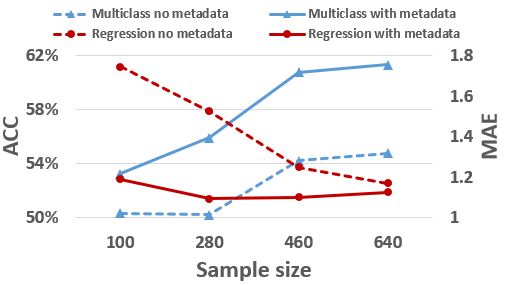}
			}\\
			\vspace*{-1em}
			\subfigure[Feature importance]{\label{ImportanceHue}
				\includegraphics[width=0.8\textwidth]{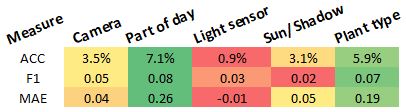}
			}
		}
	\end{minipage}
	\caption{(a) MEPP architecture. (b) Learning curves for the MEPP (solid line) and regular DNN (dashed line) in hue prediction, measured by accuracy (multiclass classification) and MAE (regression). (c) Feature importance for hue ranges from best (green) to worst (red) according to each performance measure (see Section~\ref{Results}).}
	\label{FigureArch}
\end{figure*}

Our proposed MEPP architecture is shown in Fig.~\ref{FigureArch}(a). We used the Inception V3~\citep{szegedy2016rethinking} after replacing its top layer (with 1,000 neurons for ImageNet) with a dense representation layer of 200 neurons [\underline{a}]. Each metadata variable is fed into an embedding weight matrix [\underline{b}] after it was one-hot encoded [\underline{c}]. The output of each matrix is a vector [\underline{d}] of 40/60 neurons depending on the original variable cardinality. Vectors for all metadata variables and the 200 (dense layer) neurons are then concatenated [\underline{e}] before being fully connected into another dense layer of 256 neurons [\underline{f}]. The network output is either a single neuron, when the network regresses on a single continuous photometric phenotype, or multiple neurons, when the network solves a multiclass or multitarget classification task. In the latter task, the model minimizes the loss of all targets simultaneously. 

The metadata variables are not directly connected to the network, but through embedding matrices [\underline{b}] with each transforming the variable low dimensional representation into a high one. This provides approximately the same dimension to the representations based on the metadata and topless Inception before the concatenation stage [\underline{e}]. Each embedding network [\underline{g}] consists of two fully connected layers. The first layer is responsible for the dimension transformation, and the second promotes non-linearity. For example, to transform a meta variable of dimension $a$ to size $b$ (which is approximately the dense layer dimension, 200, divided by the number of meta variables), the embedding weight matrix will be of size $a \times b$. Finally, we use log-loss (cross-entropy) and $l_2$ loss functions for discrete and continuous photometric phenotypes, respectively.  

\section{Evaluation}\label{SecEva}

First, we randomly split the dataset into training (80\%) and test (20\%) tomatoes. We split based on tomatoes and not images in order to avoid the situation in which images of the same tomato (taken, e.g., by different cameras or in different parts of the day) exist in both the training and test sets, soliciting the DNNs to learn other than the tomato photometric properties. Second, although our two target variables, hue and chroma, are continuous, their physical ranges are limited in our data, 22--31 and 37--52, respectively, and thus, in addition to consider regression, we discretized each target variable to three equal-width bins and considered also multi (three-) class classification.

To demonstrate the advantage of the MEPP, we conducted a series of experiments that compare it to an Inception V3 DNN that does not use image metadata. Four models were built for the two target variables used in regression and (multiclass) classification, and a fifth model was built for multi (two-) target classification for the simultaneous classification of hue and chroma. In the latter model, the output layer is a concatenation of both targets, e.g., for simultaneous classification of a tomato to the second hue state ([0,1,0]) and first chroma state ([1,0,0]), the concatenated output is \{0,1,0,1,0,0\}. The motivation for the multitarget (not to confuse with multilabel classification, a special case of multitarget~\citep{read2016meka}) is to utilize the dependency between the two target variables in order to predict them simultaneously.

Training of both DNNs used 200 epochs and a batch size of 20 (both were selected experimentally). Evaluation under regression is using the mean absolute error (MAE), i.e., the $L1$ loss, which gives a good sense of how far the predictions are from the true value~\citep{leon2006color}, and that under classification is using the accuracy (ACC) and F1 measures (precision and recall computation for F1 is of a class versus the other two classes, averaged over all classes). In each experiment, we used 10 data permutations and report on the average results.

\section{Results}\label{Results}

Table~\ref{Lifts} summarizes classification results for the hue (H) and chroma (C) phenotypes as multiclass (3 classes) and multitarget (6 classes) functions of the regular DNN and MEPP with respect to accuracy (ACC), F1, and lifts (the ratios between the MEPP and regular DNN) for these performance measures. The table also provides results for a Na\"ive (baseline) classifier that assigns all samples to the majority class, and since this class is represented in probabilities of 49.7\% (H) and 74.3\% (C), this approach is accurate in these probabilities. In addition, Table~\ref{Lifts} shows the performance of an expert breeder (Human) for the hue phenotype (human experts are not trained to percept chroma). This performance is evaluated by comparing the breeder decision on ''color'' to that measured by the photo-box, which provides the true label in our experiments (both decisions are one of nine levels). Table~\ref{Lifts} revels that: (1) the MEPP is always superior to the regular DNN regardless of the phenotype measured, target function, and performance measure (this is reflected in the always positive lifts). This superiority is especially vivid regarding F1, which reflects also the specific advantage of MEPP with respect to classification of the minor (hue/chroma) classes. A non-parametric Wilcoxson signed rank test~\citep{demvsar2006statistical} with a 0.05 confidence level shows that the differences in F1 between the DNN architectures regarding hue and chroma are statistically significant; 2) the MEPP is always superior to the human expert and the naive approach regardless of the phenotype measured, target function, and performance measure. A non-parametric Wilcoxson signed rank test shows this superiority is statistically significant with respect to hue and F1; and (3) while for the regular DNN, the performances achieved for the multitarget function are not always better than those for the multiclass, the MEPP multitarget performance outperforms that for multiclass, indicating better exploitation of the metadata by the MEPP when phenotyping is cast as a multitarget problem.

\begin{table}[t!]
	\centering
	\tiny
	\caption{The MEPP and regular DNN performances for the multiclass and multitarget hue (H) and chroma (C) phenotype representations. A \textbf{bold} value indicates the highest value of a measure for a phenotype.}
	\hspace*{-2em}
	\begin{tabular}{c|cccccccc|ccccccc}
		& \multicolumn{8}{c|}{ACC (\%)} & \multicolumn{7}{c}{F1} \\\hhline{~---------------}
		& \multicolumn{3}{c}{Multiclass} & \multicolumn{3}{c}{Multitarget} & \specialcellm & \specialcelln & \multicolumn{3}{c}{Multiclass} & \multicolumn{3}{c}{Multitarget} & \specialcelln \\\hhline{~------~~------~}
		& DNN & MEPP & Lift & DNN & MEPP & Lift & & & DNN & MEPP & Lift & DNN & MEPP & Lift &   \\\hline
		H & 54.7 & 59.5 & 8.8 & 55.4 & \bf{61.3} & 10.6 & 49.7 & 55.4 & 0.38 & 0.44 & 15.8 & 0.39 & \bf{0.49} & 25.6 & 0.38 \\
		C & 75.1 & 77.5 & 3.2 & 73.8 & \bf{78.0} & 5.7  & 74.3 & -    & 0.43 & 0.51 & 18.6 & 0.45 & \bf{0.52} & 15.6 & - \\
	\end{tabular}
	\label{Lifts}
\end{table}

Regarding regression, the MEPP was superior to the regular DNN for both hue (MAE of 1.07 vs. 1.42) and chroma (MAE of 1.12 vs. 1.47), whereas the human-expert error for hue was 1.20 (and unspecified for chroma, again, because it cannot be measured by humans).

Then we checked learning curves for multiclass classification and regression to examine differences with respect to speed of learning and sensitivity to the sample size between the two DNN architectures, with and without metadata. We fixed the test set (180 images), and sampled (from the remaining 640 tomato images) training sets of 100, 280, 460, and 640 images. For each training set, we trained the MEPP and regular DNNs to predict hue and chroma and tested them on the same test set of 180 tomato images. Fig.~\ref{Figurehue} shows that for hue, as the (training) sample size increases, the MAE decreases and the ACC increases. Interestingly, the ultimate MEPP MAE value is achieved already for a small sample size and is almost insensitive to the sample size, whereas, while the regular DNN deviates for small sample sizes, it makes lower deviations as the size increases, until, for a relatively large sample, the two architectures are almost similar. That is, regarding regression, the use of metadata in hue phenotyping is advantageous especially for samples that are not large. With respect to multiclass classification accuracy, both the MEPP and regular DNN architectures improve performance with the sample size, where the advantage of the former over the latter is kept almost fixed for all sizes. 

Finally, to measure the importance of each metadata variable, we exercised an ablation study, i.e., the performance was calculated where metadata variables are alternately removed. The variable, once removed, yielded the highest increase in accuracy/F1 or the highest decrease in MAE was the most important according to this measure. Fig.~\ref{ImportanceHue} shows the hue importance map with the impact of each missing metadata variable on the multiclass classification accuracy and F1 (upper two rows) and regression MAE (lower row). The value in each cell is the reduction (ACC/F1) or increase (MAE) compared to the baseline model that uses all the metadata variables. For example, 3.5\% for Camera and ACC in Fig.~\ref{ImportanceHue} stands for a 3.5\% reduction in accuracy when removing the ''Camera'' metadata variable. We can see that the results are consistent, i.e., the importance order of the variables remains the same for the three performance measures, and Part of the day and Plant type are the most important features in classifying hue in the images (although not shown, the chroma had a similar ranking).

\section{Summary and future work}

In this paper, we demonstrated that metadata embedded into a DNN improves agricultural image photometric phenotyping. Currently, this task is accomplished by human experts or in a laboratory using a dedicated photo-box that requires picking the measured fruits. To the best of our knowledge, this is the first attempt to estimate agricultural photometric phenotypes based on field-condition images.~Our MEPP DNN architecture outperformed its (Inception V3) counterpart DNN and the human expert in predicting two photometric phenotypes of tomatoes.~Another advantage of our proposed method is that MEPP can use metadata collected automatically and effortlessly, e.g., using a camera's built-in clock, open-source weather datasets, and light sensors mounted on the camera indicating on the season, day, time of the day, and cloud coverage (some or all of them), all affecting photometric phenotyping. Future research should concentrate on expanding the experiments to other crops in which photometric phenotypes are important, extending the study to additional photometric phenotypes, and developing methods to automatically extract environmental conditions, e.g., whether a fruit is in direct sun or shadow using dedicated detectors.

\section{Acknowledgment}

This work was supported by the Israel Innovation Authority through the Phenomics MAGNET Consortium.

\bibliography{iclr2020_conference}
\bibliographystyle{iclr2020_conference}

\end{document}